\documentclass{article}
\usepackage{ijcai22}

\usepackage{times}
\usepackage{soul}
\usepackage{url}
\usepackage[hidelinks]{hyperref}
\usepackage[utf8]{inputenc}
\usepackage[small]{caption}
\usepackage{graphicx}
\usepackage{amsmath}
\usepackage{amsthm}
\usepackage{booktabs}
\usepackage{algorithm}
\usepackage{algorithmic}
\usepackage{times}
\usepackage{epsfig}
\usepackage{graphicx}
\usepackage{amsmath}
\usepackage{amssymb}

\usepackage{colortbl}

\usepackage{booktabs}
\usepackage{multirow}
\usepackage{bbding}

\usepackage{hyperref}
\hypersetup{colorlinks=true,citecolor=black}

\usepackage{tabularx}
\newcolumntype{C}{>{\centering\arraybackslash}X}

\title{GL-RG: Global-Local Representation Granularity for Video Captioning}

\author{
Liqi Yan$^{1,2,8}$\footnote{Equal contributions.}
\and
Qifan Wang$^{3*}$\and
Yiming Cui$^{4}$\and
Fuli Feng$^5$\and
Xiaojun Quan$^6$ \and\\
Xiangyu Zhang$^7$ \And
Dongfang Liu$^8$\footnote{Corresponding author.}
\affiliations
$^1$Fudan University \\
$^2$Westlake University \\
$^3$Meta AI \\
$^4$University of Florida \\
$^5$University of Science and Technology of China \\
$^6$Sun Yat-sen University  \\
$^7$Purdue University \\
$^8$Rochester Institute of Technology \\
\emails
yanliqi@westlake.edu.cn,
wqfcr@fb.com,
dongfang.liu@rit.edu
}

\begin{document}

\maketitle

\begin{abstract}
  Video captioning is a challenging task as it needs to  accurately  transform  visual  understanding  into natural language description.  To date, state-of-the-art  methods  inadequately  model  global-local representation across video frames for caption generation, leaving plenty of room for improvement.  In this work, we approach the video captioning task from a new perspective and propose  a GL-RG framework for video captioning, namely a \textbf{G}lobal-\textbf{L}ocal \textbf{R}epresentation \textbf{G}ranularity. Our GL-RG demonstrates three advantages over the prior efforts: 1) we explicitly exploit extensive visual representations from different video ranges to improve linguistic expression; 2) we devise a novel global-local encoder to produce rich semantic vocabulary to obtain a descriptive granularity of video contents across frames; 3) we develop an incremental training strategy which organizes model learning in an incremental fashion to incur an optimal captioning behavior. Experimental results on the challenging MSR-VTT and MSVD datasets show that our DL-RG outperforms recent state-of-the-art methods by a significant margin. Code is available at \url{https://github.com/ylqi/GL-RG}.
  
%   For its 
% simplicity, GL-RG could serve as a strong basis for other video understanding tasks besides video captioning. 
\end{abstract}

\section{Introduction}

Video captioning is of great societal relevance, holding values for many real-world applications, including subtitle generation, blind assistance and autopilot narration. However, isolated video frames may suffer from motion blur or occlusion, which introduces great confusion in visual understanding for the captioning task. 
Therefore, there is an urgent need to answer
a principal problem: how to leverage the rich global-local features regarding to cross-frame coherence and single frame information
% \xz{as an outsider, some definition of global-local features may be helpful} 
in video contents to close the gap from visual understanding to language expression?
\begin{figure}
    \centering
    \includegraphics[width=1\linewidth]{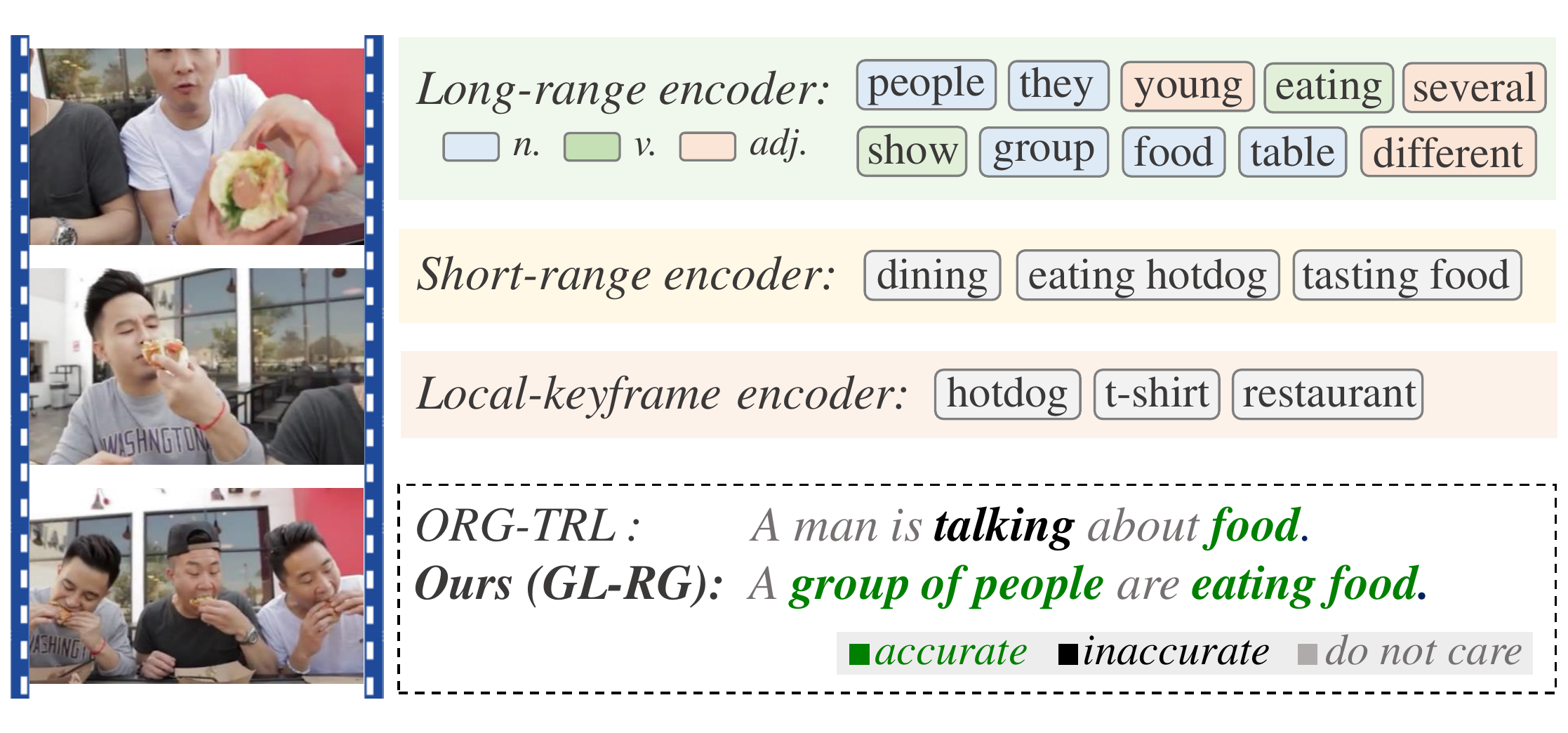}
    \caption{Qualitative examples of drastic scene variations across frames. By using global-local representations in videos, our method achieves fine-grained description of the video contents, in comparison with state-of-the-art methods ORG-TRL.} 
    \vspace{-3mm}
    \label{fig:demo}
\end{figure}

Despite making significant progress, existing methods for video captioning inadequately capture the local and global representation.
Various works apply the deep neural network on raw pixels to build higher-level connections~\cite{WangControllable,ZhangObjectAware}. These methods focus on local object features but neglecting object transformation or interaction. The effort of modeling local object features is a primitive solution for video captioning because the temporal connections across frames are not explored delicately and thus sensitive to the spurious association.
%Rather than modeling the correlations of semantic entities across frames, a plenty of works simply apply the deep neural network on raw pixels to build higher-level connections \cite{WangControllable}. The primary focus of these methods is to operate on local object features but neglecting object transformation or interaction \cite{ZhangObjectAware,Luowei2019Grounded}. The effort of modeling local object features is a primitive solution for video captioning because the temporal connections across frames are not explored delicately and thus sensitive to the spurious association.

To study the problem of the global-local correlation, other related vision tasks leverage the graph representation using graph neural networks (GNNs). For instance, \cite{TsaiVideo} models object relations by using video spatio-temporal graphs and explicitly builds links between high-level entities. %Specifically, each node encodes a target entity ($i.e.$, objects or persons \cite{FanUnderstanding}, body joints \cite{YanSpatial}, and actions \cite{TsaiVideo}), while each edge represents correlations among the entities.
Inspired by the above success, recent video captioning studies extend the graph-based approach and use GNNs to model global-local reasoning~\cite{ZhangObject,PanSpatioTemporal}. Among these works, \cite{ZhangObjectAware} merges local features with the global feature using concatenation; \cite{GhoshStacked} adds spatio-temporal features as a separate node in the graph. However, empirical results indicate that using graphs to represent global-local correlation is suboptimal as it often encounters the over-smoothing problem in training which leads to weak results during inference.
Alternatively, many video captioning methods intuitively exploit multi-modal fusion ($i.e.$, visual or audio features) to enrich the feature representation in prediction \cite{rahman2019watch}. However, these simple ``lumping'' approaches inefficiently exploit multi-modal features and struggle to perform joint optimization cross-modality, leaving large room for improvement.

% \indent In this work, we attempt to solve video captioning in a more flexible approach, which exploits the global-local vision representation for video captioning. From our global-local perspective, long-range video frames can describe spatio-temporal correspondence, short-range video frames can capture object motion and tendency, while a local keyframe can preserve finer object appearance and location details.  Collectively, the accumulative global-local representations on video frames provide sufficient information to translate visual understanding into accurate language expression.

To address the aforementioned problems, we attempt to solve video captioning in a more flexible approach, which exploits the global-local vision representation granularity. 
Concretely, we make the following contributions:
\begin{itemize}
    \item We devise a simple framework called GL-RG, namely the global-local representation granularity, which models extensive vision representations and generates rich vocabulary features based on video contents of different ranges.
    \item 
    % We devise a novel global-local encoder to generate rich vocabulary features based on the video contents of different ranges. 
    We propose a novel global-local encoder, which exploits rich temporal representation for video captioning. The encoder jointly encodes the long-range frames to describe spatio-temporal correspondence, the short-range frames to capture object motion and tendency, and the local keyframe to preserve finer object appearance and location details (Figure~\ref{fig:demo}). 
    % Collectively, the accumulative features across frames provide holistic representations to translate visual understanding into accurate language expression (Figure~\ref{fig:demo}). 
    \item We introduce an incremental two-phase training strategy. In the first seeding phase, we design a discriminative cross-entropy for non-reinforcement learning, which addresses the problem of human annotation discrepancy. In the second boosting phase, we adapt a discrepant reward for reinforcement learning, which stably estimates a bias of the expected reward for each individual video.
    % and outperforms the self-critical baseline in \cite{RennieSelf}.
    % Therefore, compared to the concurrent methods ($i.e.,$ self-critical method \cite{RennieSelf}), our network can be trained to attend to the distinction of the human-labeled annotations for each video and harmonize the contrast of all the videos themselves (Figure~\ref{fig:System}).
    % Compared to prior training methods, our training strategy can more effectively organize feature learning to obtain optimal performance for linguistic granularity.
    \item We evaluate our approach on the MSR-VTT~\cite{MSR-VTT} and MSVD \cite{MSVD} datasets. Extensive experimental results indicate that our method outperforms the latest best systems and uses shorter training schedules.
    %We also use ablative studies to verify the power of our idea and the efficacy of our algorithm.
\end{itemize}
% \begin{itemize}
%     \item We propose a simple yet novel method called \textbf{GL-RG}, which explicitly  exploit  extensive vision representations from different video ranges to improve linguistic expression. With our design, we achieve the \textbf{G}lobal-\textbf{L}ocal \textbf{R}epresentation \textbf{G}ranularity for video captioning generation, thus the name of our work.
%     \item We devise a novel global-local encoder to generate rich vocabulary features based on the video contents of different ranges (i.e., long-range, short-range, and local-keyframe), which enrich a fine-grained description of video contents across frames (see Figure~\ref{fig:demo}).
%     \item We introduce an incremental training strategy. Our training can effectively stimulate model learning in an incremental fashion to improve captioning prediction behavior. Our evaluations on MSR-VTT~\cite{MSR-VTT} and MSVD \cite{MSVD} dataset demonstrate state-of-the-art performance.
%     % \item Evaluated on MSR-VTT~\cite{MSR-VTT} and MSVD \cite{MSVD} dataset, our method outperforms the latest best systems with shorter training schedules. 
% \end{itemize}

\section{Related Work}
% \xz{Overall, I see that you criticize existing works a lot. I am in general not in favor of such a method because the authors of these works are likely the ones that review your paper. One way to spin it would be to say good things about existing ones and say yours is complementary. For example, your training strategy can be considered complementary to the existing ones. You don't have to compete with them} 
\textbf{Video captioning.} Inspired by the success of other vision tasks, the seminal  work \cite{venugopalan2015sequence} extends the encoder-decoder architecture for the video captioning task. Following the same paradigm, \cite{chen2018less,venugopalan2015sequence} explore the temporal patterns on video using attention mechanisms to depict object movements. \cite{pei2019memory} devises a MARN method, which generalizes descriptions from a single video to other videos with high semantic similarity. \cite{HouJoint} develops an idea of feature fusion to guide sentence generation for video content. Different from the existing efforts, we explicitly explore global-local representations for sentence generation.
\begin{figure*}[tbp]
    \centering
    \includegraphics[width=0.9\linewidth]{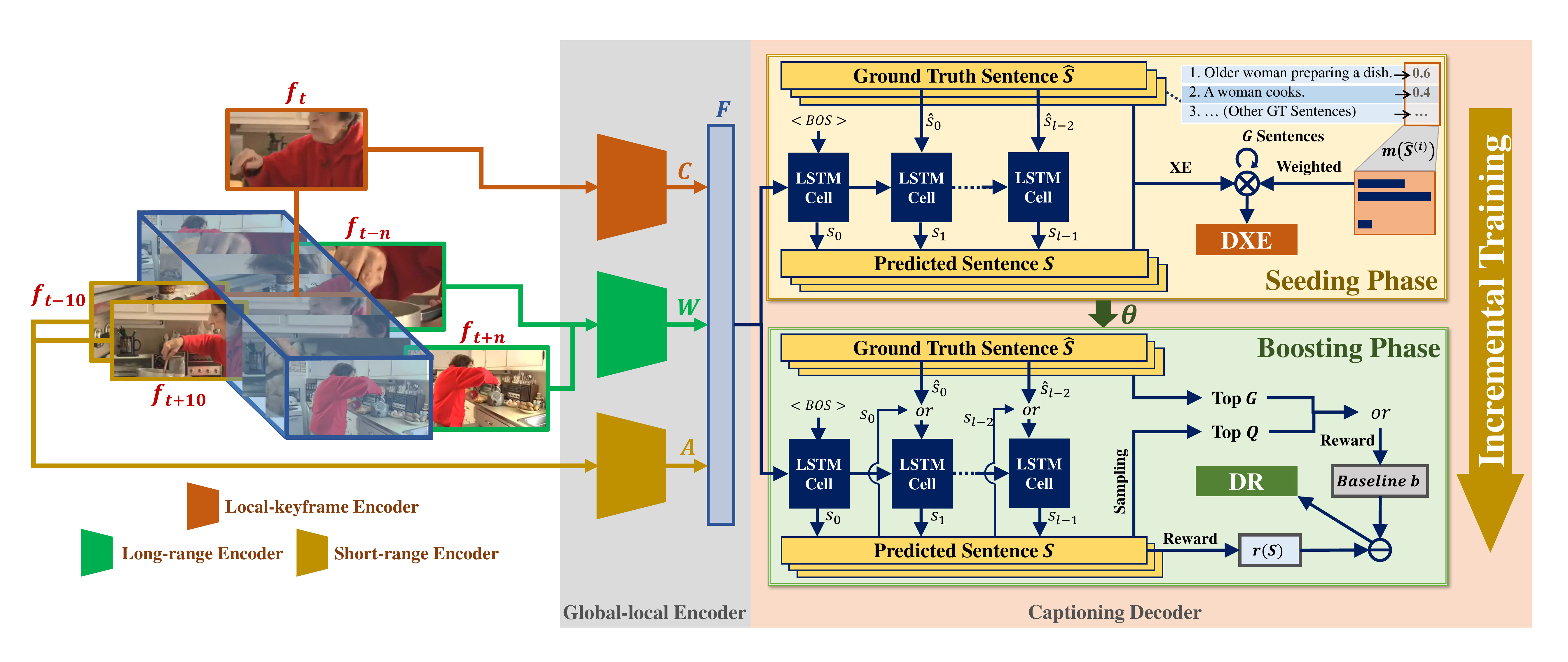}
    \caption{The architectural framework of GL-RG. Our global-local representation encoder includes: 1). the long-range encoder captures temporal correspondence
    among distant frames ($t-n$ to $t+n$ frames) and makes the cross-frame
    representations robust to appearance variations and shape deformations; 2). the short-range encoder focuses on motion and tendency, which depicts the local consistency of object movement within a short moment ($t\pm10$ frames); 3). the local-keyframe encoder focuses on each object, which can preserve better object spatial information and finer details in terms of object appearances. In training, our method is trained by an incremental strategy which includes a seeding phase and then a boosting phase. The seeding phase supervises our method to obtain an entrance model which can be easily trained in the second boosting phase. }
    \vspace{-3mm}
    \label{fig:System}
\end{figure*}\\
\textbf{Global-local representation.} To model the global-local vision representation, many methods \cite{liu2021sg,Zhang_2021_CVPR,liu2020video} resort to the sequence learning strategy. \cite{yao2015describing} uses a temporal attention method to depict the global-local connections. \cite{WangControllable} leverages the decoding hidden states to increase the temporal feature representation. More recently, \cite{hu2019hierarchical,yang2017catching,ZhangObjectAware} exploit the object features to model the object movement across frames. For instance, \cite{ZhangObjectAware} employs a bidirectional temporal graph to capture detailed movements for the salient objects in the video; \cite{hu2019hierarchical} devises a stacked LSTM to encode both the frame-level and object-level temporal information. However, the above work primarily focuses on feature salience from the global contents with less consideration of the global-local representation reasoning. In contrast, we model global-local representations to achieve the lexical granularity, using long-range temporal correspondence, short-range object motion, and local spatial appearances on video contents.\\ 
% With the global-local mechanism, our method achieves fine-grained language expression for video captioning.\\
\textbf{Training strategies.} A popular strategy for training video captioning models is “Teacher Forcing” \cite{williams1989learning}, which has been widely used in training video captioning tasks \cite{ZhangBridging}. More recently, many research efforts attempt to explore different training methods to boost captioning performance \cite{wang2018video,RennieSelf,HouJoint,Ryu2021SemanticGN}. For instance,  \cite{pasunuru2017reinforced} uses a mixed loss function to optimize the video captioning algorithm, which leverages the weighted combination of cross-entropy and reinforcement learning. Similarly, \cite{RennieSelf} adopts the paradigm of reinforcement learning and devises a self-critical baseline to reward the model learning to train the video captioning network. Although demonstrating appealing supervision performance \cite{Deng_2021_CVPR,Fenglin_2021_ACL}, the above methods generally require a complicated pipeline to train with a computation overhead for optimization. Building on the lessons learned from the concurrent approaches, we propose an incremental training strategy, which can easily operate training on our proposed GL-RG. Empirical results indicate that our training strategy can serve as a good addition to the existing training scheme to boost a further training gain.
\section{GL-RG}
\subsection{Overview}
The framework of GL-RG is demonstrated in Figure ~\ref{fig:System}. Following~\cite{PanSpatioTemporal}, GL-RG also adopts an encoder-decoder architecture. More specifically, we include a global-local encoder and a captioning decoder. The global-local encoder selects frames of different ranges as inputs and encodes them into different vocabulary features. All the obtained features are aggregated together to enrich global-local vision representations across video frames. Afterwards, the captioning decoder supervised by the incremental training strategy translates the vocabulary feature into natural language sentences. We elaborate on the proposed GL-RG below.
\subsection{Global-Local Encoder}
\label{sec:encoder}
Our global-local encoder includes three essential parts: long-range encoder, short-range encoder, and keyframe encoder (see Figure ~\ref{fig:System}). Collectively, our encoder can enrich global-local vision representations for video captioning tasks.\\
\textbf{Long-range encoder.} We encode random global video frames to produce the global vocabulary based on a random keyframe $f_t$ in training.
% \xz{how do you select the random frames? Are the results sensitive to your random selection? How many do you select? Here, there is ambiguity, are you selecting all frames in a random range or random frames in a fixed range?}
Note that our training iterations will fully saturate the whole video clips, since each iteration will randomly choose different frames (the total number is fixed) from the videos.
% \textcolor{magenta}{(We have 32 keyframes (explained in the experiment section). We select 2n frames and $f_t$ is based on random seed in training.)}
Our long-range encoder first performs 2D convolutions on the inputs ($i.e.$, $f_{t-n}$ and $f_{t+n}$\footnote{where $n$ is a random range larger than 25 frames.}) to identify the relevant contextual features. The output features from the first step are processed by a 3D convolutional network (CNN) to capture global temporal correspondence. In order to increase consensus, we choose the top $K$ word choices (highest frequency) from the ground truth sentences to guide the vocabulary generation as a $K$-classification task. Outputs of the dense layers are defined as:
\begin{equation}
    W = \{w_{1},w_{2},...,w_{k}...,w_{K}\}, w_{k} \in (0,1)
\end{equation}
% \xz{Is your notation here correct? With the current form, W may have a dynamic size $j$. Do you mean W is always size $K$?}
% \textcolor{magenta}{(The top $K$ words are the highest frequency of K words from all annotated sentences of all videos in the whole video captioning dataset, so $K$ is a fixed number; For a video, $W$ is the vector is composed of probabilities that each word will be used to describe the video contents.)}
where $W$ is the collection of predicted  long-range vocabulary. Since  $W$ includes all possibilities of word choices from the ground truths ($i.e.$, MSR-VTT \cite{MSR-VTT} and MSVD \cite{MSVD}), it can offer the description of temporal contents in video. 
It is consisted of top $K$ frequently used words (such as verbs, nouns and adjectives, excluding ``is", ``be" and ``do", ext.) from all annotated GT sentences of all videos in the dataseta, $i.e.$, MSR-VTT and MSVD.\\
% \xz{what is long-range vocabulary? are these words only related to temporal contents? From your description, your ground truth are just top K words, which may not be temporal at all. I guess you miss a clear definition}
% \begin{figure}
%     \centering
%     \includegraphics[width=0.9\linewidth]{IMG/long_range.pdf}
%     \caption{The architecture of the long-range encoder.}
%     \label{fig:long-range}
% \end{figure}
\textbf{Short-range encoder.}  Our short-range encoder is to capture object motion and tendency.
Specifically, simultaneously taking two close neighbours (a.k.a. $f_{t-10}$ and $f_{t+10}$) of the keyframe, 2D CNN and 3D-Resnet18 \cite{TranConvNet} yield the semantic and movement representations respectively. Afterwards, these representations are stacked and fed into dense layers for action classifications. Given the highest frequency of $J$ actions (in Kinetics-400, UCF101 or HMDB datasets), our output of the short-range encoder is the following.
\begin{equation}
    A = \{a_{1},a_{2},...a_{j},...a_{J}\}, a_{j} \in (0,1),  
\end{equation}
% \xz{I have the same notation question here}
% \textcolor{magenta}{($J$ is the number of all actions in video action recognition task. For one frame sequence, $A$ is the list of all the confidences of those $J$ actions predicted by the model.)}
% \xz{still, your notation seems to be wrong. What it means is that A is a vector with various sizes from 1 to J. Sometimes, it may be size 1, some other times it may be size J. Is that what you mean?}
% \textcolor{magenta}{(Sorry, there the size of A is fixed to J, I modified the notation now.)}
where $A$ is the collection of $a_{j}$,
% \xz{where do you get this vocabulary?} \textcolor{magenta}{(The actions are from the video action recognition dataset, such as Kinetics-400, UCF101, HMDB.)} 
which is the predicted confidence of the $j_{th}$ action from the short-range action dataset.\\
% \begin{figure}
%     \centering
%     \includegraphics[width=0.9\linewidth]{IMG/short-range.pdf}
%     \caption{The architecture of the short-range encoder.}
%     \label{fig:short-range}
% \end{figure}
\textbf{Local-keyframe encoder.} The lexical knowledge for the local semantics is learned by a residual network~\cite{XieAggregated}, which extracts salient object features from the keyframe $f_{t}$. Given the number of image classes in the image classification dataset (e.g. ImageNet) is $M$, the output of our local encoder is:
\begin{equation}
    C = \{c_{1},c_{2},...,c_{m},...,c_{M}\}, c_{m} \in (0,1),
\end{equation}
% \xz{is notation correct?}
% \textcolor{magenta}{(It's the same as the two encoders before.)}
where $C$ is the collection of $c_{m}$, which is
% \xz{what is this vocabulary and how do you define it?} 
% \textcolor{magenta}{(the same with the dataset of image classification dataset, such as ImageNet.)} 
the predicted confidence of the $m_{th}$ class for this local frame.\\
\indent Once having all the vocabulary features from different ranges, we perform a fusion encoding. We first use a feature pool composed of linear layers $\varphi$ to project each vocabulary feature into a same-size embedding and then aggregate them together to produce the fused feature $F$:
\begin{equation}
    F = \mathrm{Concat} (\varphi (W), \varphi (A), \varphi (C))
\end{equation}
\subsection{Captioning Decoder}
% \textbf{Sequence-to-Sequence decoding.} 
Our captioning decoder translates the fused feature into a $ l $-word\footnote{$l$ ($i.e.$, $30$ in our experiment) denotes the maximum length of a sentence.} sequence $S=(s_1, s_2,..., s_j|j\in\{1,...,l\}\}$
% \xz{wrong notation to represent a $l$-word sequence} 
% \textcolor{magenta}{(See the footnote, the number of the word can be smaller than $l$, since $l$ is the maximum length of a sentence.)} 
to form the predicted sentence. Specifically, we use a LSTM to generate a hidden state $ h_t $ and a cell state $ c_t $ at the $ i_{th} $ step:
\begin{equation}
    \begin{aligned}
    h_{i}, c_{i} = \mathbb{LSTM}([h_{i-1}, \Phi (s_{i-1}, \hat{s}_{i-1}, F)], c_{i-1}),
        \label{asds}
    \end{aligned}
\end{equation}
where $ [\cdot,\cdot] $ denotes concatenation. $ h_{i-1}, s_{i-1}, \hat{s}_{i-1}, F$, and $c_{i-1}$ are the previous hidden state, the predicted word, the ground truth, the fused feature from encoding, and the cell state respectively. $\Phi(\cdot)$ is the annealing scheme which uses every previous token to predict the next word. We adopt a schedule sampling technique to randomly choose the token $s_{i-1}$ or $\hat{s}_{i-1}$ using a random variable $\xi \in \{0, 1\}$:
\begin{equation}
    \begin{aligned}
    \Phi (s_{i-1}, \hat{s}_{i-1}, F) = 
    \begin{cases} 
    F, & (i=1); \\
    s_{i-1}, & (i>1, \xi=0); \\
    \hat{s}_{i-1}, & (i>1, \xi=1),
    \end{cases}
    \end{aligned}
    \label{666666}
\end{equation}
when $i=0$, the initial input of the LSTM is the fused feature $F$; when $i>1$, we increase the probability of $\xi=1$ gradually in every epoch until $\xi$ is absolutely equal to 1. Then, we counter the process by decreasing the probability of $\xi=1$.

Accordingly, the probability of a predicted word is:
\begin{equation}
    \begin{aligned}
p_{\theta}(s_i|h_i)=\mathrm{softmax}(W_o\cdot h_i),
    \end{aligned}
\end{equation}
where $h_i$ is the hidden state from Eq.~(\ref{asds}) and $W_o$ is the weight matrix %\cite{WisemanBeam-Search}
which maps the hidden state $h_i$ to vocabulary-sized embedding, in order to find a context-matching word in the sentence. 
\begin{table*}[tb]
\centering
\scalebox{0.83}{\begin{tabular}{c|c|c|ccc|cccc|cccc}
    \toprule 
     \multirow{2}{*}{Training} & \multirow{2}{*}{Method}     & \multirow{2}{*}{Epoch}     & \multicolumn{3}{c|}{Feature}  & \multicolumn{4}{c|}{MSR-VTT} & \multicolumn{4}{c}{MSVD}  \\
    \cline{4-6}
   \rule{0pt}{12pt} &            &          & Local   & Short   & Long & B@4 & M  & R  & C  & B@4  & M  & R  & C  \\
    \midrule
    \multirow{10}{*}{XE} 
    & SA-LSTM \cite{MSR-VTT}               & 100      & \checkmark  & \checkmark  & $\times$ & 36.3 & 25.5 & 58.3 & 39.9   & 45.3 & 31.9 & 64.2 & 76.2     \\
	& RecNet \cite{wang2018reconstruction} & -       & \checkmark & $\times$  & $\times$ & 39.1    & 26.6  & 59.3  & 42.7 & 52.3 & 34.1 & 69.8 & 80.3    \\
	& ORG-TRL \cite{ZhangObject}           & -    & \checkmark   & \checkmark   & $\times$ & 43.6      & 29.7     & 62.1      & 50.9  & 54.3 & 36.4 & 73.9 & 95.2   \\
	
	& STGraph \cite{PanSpatioTemporal}     & 50   & \checkmark      & $\times$  & $\times$   & 40.5      & 28.3     & 60.9      & 47.1 & 52.2 & 36.9 & 73.9  & 93.0    \\
	& SGN \cite{Ryu2021SemanticGN}         & -    & \checkmark      & $\times$  & $\times$   & 40.8    & 28.3     & 60.8       & 49.5 & 52.8 & 35.5 & 72.9  & 94.3 \\
	& O2NA \cite{Fenglin_2021_ACL}         & 50   & \checkmark   & \checkmark   & $\times$  & 41.6    & 28.5   & 62.4    & 51.1 & 55.4    & 37.4  & 74.5    & \textcolor{red}{96.4} \\
	& RCG \cite{Zhang_2021_CVPR}           & -     & \checkmark   & \checkmark   & $\times$    & 42.8    & 29.3   & 61.7  & \textcolor{blue}{52.9}  & - & - & - & -  \\
	
    % \midrule
    
    & \textbf{Ours (GL-RG)}             
	      & \textbf{30}   & \checkmark  & \checkmark    & \checkmark & \textbf{\textcolor{blue}{45.5}}  & \textbf{\textcolor{blue}{30.1}}  & \textbf{\textcolor{blue}{62.6}}  & \textbf{51.2} & \textbf{\textcolor{blue}{55.5}} & \textbf{\textcolor{blue}{37.8}} & \textbf{\textcolor{blue}{74.7}} & \textbf{94.3}  \\
	   
	\midrule
	% \cline{1-1}
	\textbf{DXE}
	& \textbf{Ours (GL-RG)}             
	        & \textbf{30}   & \checkmark & \checkmark   & \checkmark   
	    & \textbf{\textcolor{red}{46.9}}  & \textbf{\textcolor{red}{30.4}}  & \textbf{\textcolor{red}{63.9}}   & \textbf{\textcolor{red}{55.0}}   & \textbf{\textcolor{red}{57.7}}      & \textbf{\textcolor{red}{38.6}}    & \textbf{\textcolor{red}{74.9}}      & \textbf{\textcolor{blue}{95.9}}  \\
    
    \midrule
    \multirow{5}{*}{RL} 
    & HRL \cite{wang2018video}   & -       & \checkmark    & $\times$  & $\times$ & 41.3   & \textcolor{blue}{28.7}  & 61.7   & 48.0  & - & - & - & -  \\
    & PickNet \cite{chen2018less}           & 300     & \checkmark   & $\times$  & $\times$ & 38.9   & 27.2 & 59.5    & 42.1   & 46.1   & 33.1    & 69.2   & 76.0  \\
    & POS$_{RL}$ \cite{WangControllable}    & -       & \checkmark    & \checkmark  & $\times$ & 41.3     & \textcolor{blue}{28.7}     & \textcolor{blue}{62.1}      & \textcolor{blue}{53.4}  & \textcolor{blue}{53.9}   & \textcolor{blue}{34.9}    & \textcolor{blue}{72.1}   & \textcolor{blue}{91.0}   \\
    & VRE \cite{ShiWatch}               & -      & \checkmark   & $\times$  & $\times$ & \textcolor{blue}{43.2}     & 28.0      & 62.0        & 48.3  & 51.7   & 34.3    & 71.9   & 86.7 \\
    & SAAT$_{RL}$ \cite{ZhengSyntaxAware}    & 200    & \checkmark  & \checkmark  & $\times$  & 39.9      & 27.7     & 61.2      & 51.0   &  46.5 & 33.5 & 69.4 & 81.0  \\
    
    % \cline{1-1}
    \midrule
	\textbf{RL+DR}
	& \textbf{Ours (GL-RG + IT) }          
	      & \textbf{100}   & \checkmark     & \checkmark   & \checkmark   
	    & \textbf{\textcolor{red}{46.9}}   & \textbf{\textcolor{red}{31.2}}  & \textbf{\textcolor{red}{65.7}}   & \textbf{\textcolor{red}{60.6}}   & \textbf{\textcolor{red}{60.5}}      & \textbf{\textcolor{red}{38.9}}    & \textbf{\textcolor{red}{76.4}}      & \textbf{\textcolor{red}{101.0}}  \\
	\bottomrule
\end{tabular}}
\caption{Comparisons with state-of-the-art methods on MSR-VTT and MSVD datasets. The \textcolor{red}{best} and the \textcolor{blue}{second-best} methods are highlighted. In the training column, ``XE'' is cross-entropy; ``DXE'' is discriminative cross-entropy; ``RL'' is reinforcement learning; ``DR'' is the discrepant reward. ``Epoch'' indicates the training schedule for each compared method. ``L'', ``S'', and ``L'' in ``Feature'' indicate the local, short, and long visual representation. ``IT'' in our method stands for incremental training, which optimizes the CIDEr metric in boosting phase. 
}
\label{table:results}
\end{table*}

\subsection{Incremental Training}
Our incremental training includes a seeding phase followed by a boosting phase (see Figure ~\ref{fig:System}). The two training phases fulfill different learning objectives. The seeding phase aims to produce an entrance model to facilitate smooth training in the second phase, while the boosting phase leverages reinforcement learning (RL) to boost the performance gain.\\
% We propose an incremental training strategy to fulfill different learning objectives. Our incremental training includes a seeding phase followed by a boosting phase (see Figure ~\ref{fig:System}). 
% In the seeding phase, our learning is optimized by cross-entropy, which produces an entrance model to facilitate smooth training in the second phase. In the boosting phase, our training leverages reinforcement learning (RL) to achieve further performance gain.\\
% \xz{it is unclear if you pretrain the encoders and decoder first or train all of them together in one shot. Note that when you discuss encoders, you use vocabularies, which seem to suggest that you pretrain them} 
% \textcolor{magenta}{(Our encoders are pretrained firstly. We have the discussion in Experiments section.)}
\textbf{Seeding phase.} 
The existing models \cite{PeiMemoryAttended,ZhangObjectAware,ZhangObject} are commonly trained with the cross-entropy (XE) loss, which measures the average similarity of the generated sentence and all the ground truth sentences.
Since different annotators may interpret video content differently, the ground truth from the training dataset may include annotation bias. \textit{We argue that direct comparison between the captioning predictions to the ground truths cannot yield the optimal training outcomes}. We thus employ the metric scores $ m(\hat{S})$ of all ground truths
% \xz{need explanation} 
% \textcolor{magenta}{(Manually annotated ground truths have severe bias, some ground truth sentences are well-written, while others are not well-written (ambiguous or inappropriate). Thus, we use metric scores as weights, to make our training biased towards those well-written ground truth sentences.)}
% which causes the issue of overfitting in training. Thus, we use weight to balance out the probability of outfitting when
% Every video has a lot of artificially labeled sentences in the video captioning dataset, there are some good sentences as well as bad sentences, and the evaluation 
% metrics (frequently used in Natural Language Processing, NLP) should be used to get the score of each artificially labeled sentence to measure its referential importance (value), to be the weight when use this sentence to computing the cross-entropy loss with the predicted sentence.)}
as a discriminative weight in computing cross-entropy to make our training biased towards those well-written ground truths.\\ \indent Understandably, manually annotated ground truths have severe bias, that is, some ground truth sentences are well-written, while others are ambiguous or inappropriate. Metric scores encourage the training to focus on the well-written sentences. The $ m(\hat{S})$ can use different options, such as BLEU\_4, METEOR, ROUGE\_L \cite{metric}, and CIDEr \cite{VedantamCIDERr}. The analysis of each option will be reported in the experiments.

\indent Providing each video is annotated by $ G $ sentences 
% $ \mathbb{\hat{S}}=\{\hat{S}^{(1)}, \hat{S}^{(2)},..., \hat{S}^{(j)}|j\in \{1,..., G\}\}  $
$ \mathbb{\hat{S}}=\{\hat{S}^{(1)}, \hat{S}^{(2)},..., \hat{S}^{(G)}\}  $, the \textit{discriminative cross-entropy (DXE)} loss function is:
\begin{equation}
    L_{DXE}(\theta) = - \frac{1}{G} \sum_{j=1}^{G} m(\hat{S}^{(j)})\mathrm{log}\ p(\hat{S}^{(j)} | F; \theta),
\end{equation}
where $m()$ can be deemed as a constant, computed with every ground truth of each sentence. Our DXE loss increases the probability of generating captions with a high metric score by assigning higher weights to well-written ground-truth sentences. The gradient of DXE is calculated by the weighted difference between the prediction and all the target descriptions. Consequently, we DXE encourages feature learning which increases the probability of generating captions with a high metric score. 
% \xz{two questions: are you considering $m()$ as differential functions or just constants? If it is the latter case, since $m()$ changes a lot with the model output and you consider them as constant weights, the training may have difficulties converging. Is a similar training method being used by others? Second, it seems you are not normalizing these weights. As a result, your training may be biased towards those that can have high weights.}
% \textcolor{magenta}{(The result of m() is considered as a constant in our loss function, every GT sentence has a different computed value. Theoretically, we want to encourage predictions with higher scores. Empirical results resonate with our assumption.)}
The result of $m()$ is considered a constant in our loss function. Every GT sentence has a different computed value. Intuitively, we want to encourage predictions with higher scores. Empirical results resonate with our assumption.

Different from the weighted loss entropy (which manually assign weights to all categories to address the problem of unbalanced data), the weight $ m(\hat{S}) $ of our DXE is automatically calculated through metrics, evaluating the quality among all annotations. Our DXE assigns higher weights to high-quality annotations, helps the model generate captions closer to them.\\
\textbf{Boosting phase.} After the seeding phase, we employ reinforcement learning %\cite{WilliamsSimple}
with a \textit{discrepant reward (DR)} to further boost the performance of our GL-RD model. To optimize the model parameters $ \theta $, conventional methods using reinforcement learning performs a non differentiable reward in training:
\begin{equation}
    \begin{aligned}
    \nabla_{\theta} L_{DR}(\theta) &= - \sum_{S} r(S) \nabla_{\theta} p(S | F; \theta)\   \\
    &= - \mathbb{E}_{S\sim p} [r(S) \nabla_{\theta} \mathrm{log} p(S | F; \theta)],
    \end{aligned}
\end{equation}
where $\mathbb{E}_{S\sim p}$ denotes the expected value of the distribution, the reward $ r(S) $ is the evaluation metric score of the sampled sentence, and $F$ is the fused feature extracted from our global-local encoder. One problem with this training strategy is that the reward function $ r(S) $ is always positive because the metric score ranges between 0 and 1. Therefore, we can only encourage feature representations in learning but cannot perform suppression.

To address this issue, our DR is equal to the original reward $ r(S) $ subtracts a bias $b$, which is \textit{baseline}. With the bias term, our learning can be more robust to variation in prediction. Then the policy gradient can be defined as:
\begin{equation}
    \begin{aligned}
    \nabla_{\theta} L_{DR}(\theta) = - \mathbb{E}_{S\sim p} [(r(S) - b) \nabla_{\theta} \mathrm{log} p(S | F; \theta)],
    \end{aligned}
    \label{eq:reward_with_baseline}
\end{equation}
where $ b \approx E[r(S)] $. Previous self-critical method SCST \cite{RennieSelf} utilizes the reward of the greedy output at the test time as the baseline $b_{scst}$, at the cost to run inference again in every training iteration. In our implementation, the baseline $b$ has two variants: 1. $b_1$ obtained by the $ G $ ground-truth captions; and 2. $b_2$ the top $ Q $ sentences sampled by the model with the highest score during forward step. Note that either baseline ($b_1$ or $b_2$) can reduce the variance of the gradient without changing the expected value of gradient because of $\sum_{S} b \nabla_{\theta} p(S | F; \theta) = 0$. When updating gradients, this gradient $ \nabla_{\theta} $ can be approximated by Monte-Carlo sampling through a single training example. So the final gradient of our discrepant reward is:
\begin{equation}
    \begin{aligned}
    \nabla_{\theta} L_{DR}(\theta) \approx - (r(S) - r(S^{b_j})) \nabla_{\theta} \mathrm{log} p(S | F; \theta),
    \end{aligned}
\end{equation}
where $S^{b_j}$ can be used by either baseline ($b_1$ or $b_2$). In our experiments, we  carry out ablation studies to discover the impact of $b_1$ and $b_2$ on the captioning performance.

\section{Experiments}

\subsection{Implementation Details}
\indent \textbf{Dataset.} 
We evaluate our GL-RG on MSR-VTT dataset \cite{MSR-VTT}. Each video is associated with 20 ground-truth captions given by different workers. We follow the data split of 6513 videos for training, 497 videos for validation, and 2990 videos for testing. We also evaluate our GL-RG on the MSVD dataset \cite{MSVD}. We split the dataset into a 1,200 training set, 100 validation set, and 670 testing set by the contiguous index number.\\ 
\textbf{Evaluation Metrics.} We evaluate our method on four commonly used metrics BLEU\_4, METEOR, ROUGE\_L \cite{metric}, and CIDEr \cite{VedantamCIDERr}, which
% and SPICE \cite{AndersonSPICE}, 
are denoted as \texttt{B@4}, \texttt{M}, \texttt{R}, and \texttt{C} respectively.\\
\textbf{Training setup.} Our \textit{long-range encoder} is pre-trained on the video-to-words dataset ($K$=300 words) extracted from MSR-VTT or MSVD. Our \textit{short-range encoder} is pre-trained on Kinetics-400 dataset \cite{Carreira2017QuoVA}, which includes $J$=400 actions. Our \textit{Local-keyframe encoder} is pre-trained on ImageNet, which includes $M$=1000 objects. Our decoder is trained with the learning rate of 0.0003 in \textit{the seeding phase}, and 0.0001 in \textit{the boosting phase}. For each video, training is operated on 20 or 17 ground-truth captions for MSR-VTT or MSVD respectively.
\begin{figure*}
    \centering
    \includegraphics[width=0.95\linewidth]{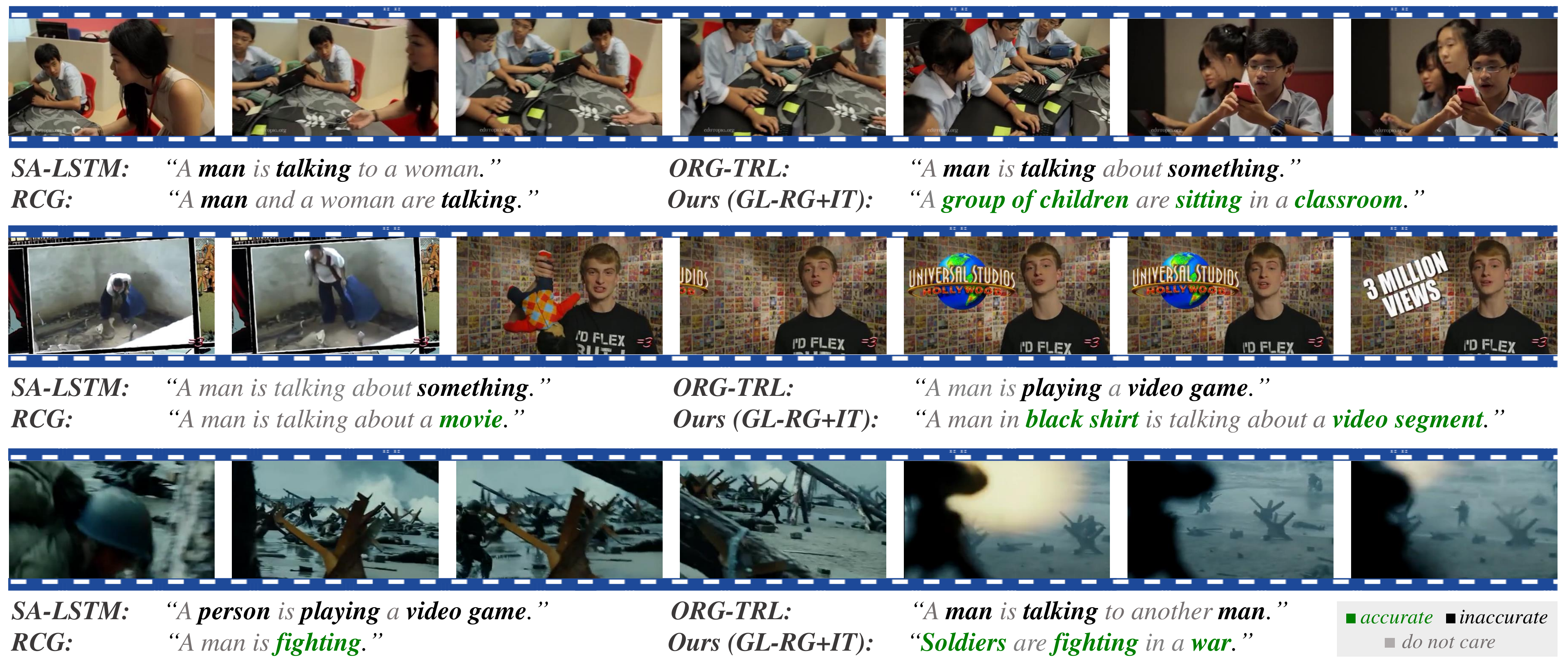}
    \caption{Qualitative results on MSR-VTT. We present comparisons with state-of-the-art methods SA-LSTM, ORG-TRL, and RCG.}
    \label{fig:Result_Compare}
\end{figure*}
\begin{table}[tb]
\centering
\scalebox{0.83}{
\begin{tabular}{p{21pt}<{\centering}p{21pt}<{\centering}p{21pt}<{\centering}|cccc}
    \toprule 
    \multicolumn{3}{c|}{Features} & \multirow{2}{*}{B@4}  & \multirow{2}{*}{M} & \multirow{2}{*}{R} & \multirow{2}{*}{C}  \\
    \cline{1-3}
    \rule{0pt}{10pt}Local & Short & Long &   &  & \\
    \midrule
	\multicolumn{7}{c}{Single feature} \\
    \midrule
	\checkmark & $\times$   & $\times$ & 31.7 & 24.0  & 54.5 & 35.3  \\
	$\times$   & \checkmark & $\times$ & 32.3 & 23.9 & 54.1 & 34.8   \\
	$\times$   & $\times$   & \checkmark & 43.8 & 28.7  & 61.2 & 51.7  \\
	\midrule
	\multicolumn{7}{c}{Combined features} \\
	\midrule
	\checkmark & \checkmark & $\times$ & 36.7$\uparrow_{5.0}$ & 26.0$\uparrow_{2.0}$  & 57.9$\uparrow_{3.4}$ & 42.3$\uparrow_{7.0}$  \\
	\checkmark & $\times$   & \checkmark & 45.1$\uparrow_{13.4}$ & 29.3$\uparrow_{5.3}$  & 62.0$\uparrow_{7.5}$ & 53.0$\uparrow_{17.7}$  \\
	$\times$   & \checkmark & \checkmark & 45.6$\uparrow_{13.9}$ & 29.3$\uparrow_{5.3}$ & 62.9$\uparrow_{8.4}$  & 53.9$\uparrow_{18.6}$  \\
	\checkmark & \checkmark & \checkmark & \textbf{46.9$\uparrow_{15.2}$} & \textbf{30.4$\uparrow_{6.4}$} & \textbf{63.9$\uparrow_{9.4}$} & \textbf{55.0$\uparrow_{19.7}$}  \\
	\bottomrule
\end{tabular}}\\
\caption{Comparison of using different features (local, short, and long-range features) in the seeding phase. DXE loss is used. $\uparrow$ is the increase from the baselines which use single feature.}
\vspace{-3mm}
\label{1111}
\end{table}
\subsection{Comparison with State-of-the-Arts}
% \xz{this should come before ablation analysis. Some case studies, e.g., with interpretation techniques applied, may be helpful, especially showing the synergy of local and global features. You can put them in appendix}
The evaluation results on the MSR-VTT dataset are shown in Table~\ref{table:results}. With a much shorter training schedule, we can achieve an on-par performance with other state-of-the-art methods. Our fully trained model also surpasses all the compared methods on all metrics. In addition, when using the same level of supervision, our margins (model trained by XE) over the next best method (ORG-TRL \cite{ZhangObject}) are 1.9\% on \texttt{B@4}, 0.4\% on \texttt{M}, 0.5\% on \texttt{R}, and  0.3\% on \texttt{C} respectively. We can achieve further performance gain by using DXE on \texttt{M}, \texttt{R}, and \texttt{C} metrics. We further compare our GL-RG against some of the recent leading methods on the MSVD dataset (see Table~\ref{table:results}). When trained by DXE, our margins over the next best method (O2NA \cite{Fenglin_2021_ACL}) are 2.3\% on \texttt{B@4}, 1.2\% on \texttt{M} and 0.4\% on \texttt{R} respectively. It is worth mentioning that our results of XE and DXE are from the $30^{th}$ epoch (as seeding phase). It can be seen that they not only outperform those latest best systems, but also use shorter training schedules. Note that all the previous RL methods are trained with self-critical baseline $b_{scst}$. Figure ~\ref{fig:Result_Compare} demonstrates some qualitative examples. 
% Compared to other methods, our method achieves improved captioning behavior.
\subsection{Ablation Study}
In this section, we conduct extensive ablation studies to analyze the effects of configurable components in GL-RG.
\begin{table}[tb]
\centering
\scalebox{0.83}{\begin{tabular}{c|c|cccc}
    \toprule 
    & $ m(\hat{S}^{(i)}) $
        & B@4     & M  & R  & C    \\
	\midrule
	XE & -   & 45.5  & 30.1  & 62.6  & 51.2 \\    
	\midrule
	\multirow{4}{*}{DXE} 
	& B@4 & 45.5$\uparrow_{0}$  & 29.7$\downarrow_{0.4}$  & 63.0$\uparrow_{0.4}$  & 51.4$\uparrow_{0.2}$ \\
	& M   & 44.5$\downarrow_{1.0}$  & 29.8$\downarrow_{0.3}$   & 62.9$\uparrow_{0.3}$  & 52.4$\uparrow_{1.2}$ \\
	& R   & 45.2$\downarrow_{0.3}$  & 29.0$\downarrow_{1.1}$  & 63.5$\uparrow_{0.9}$  & 52.5$\uparrow_{1.3}$ \\
	& C   & \textbf{46.9$\uparrow_{1.4}$}  & \textbf{30.4$\uparrow_{0.3}$}  & \textbf{63.9$\uparrow_{1.3}$}   & \textbf{55.0$\uparrow_{3.8}$}  \\
	\bottomrule
\end{tabular}}
\caption{Comparison of using different weighting metric $ m(\hat{S}^{(i)}) $ for DXE in the seeding phase. $\uparrow$ and $\downarrow$ indicates the performance change from the methods trained by XE.}
\label{table:DXE_results}
\end{table}
\begin{table}[tb]
\centering
\scalebox{0.83}{\begin{tabular}{c|ccc}
    \toprule 
    & Start From   & R  & C    \\
	\midrule
	Seeding Phase& -   & 62.6  & 51.2  \\    
	\midrule
	\multirow{2}{*}{Boosting Phase} 
	& XE  & 63.3$\uparrow_{0.7}$  & 55.3$\uparrow_{4.1}$ \\
	& DXE            
	    & \textbf{65.7$\uparrow_{3.1}$}   & \textbf{60.6$\uparrow_{9.4}$}  \\
	\bottomrule
\end{tabular}}
\caption{Comparison of using XE or DXE-trained entrance model in the boosting phase. $\uparrow$ indicates the increase from the seeding phase.}
\label{table:germinating_from_XE_DXE}
\end{table}\\
\textbf{Global-local features.} We measure the performances of our model using different global-local features (see in Table~\ref{1111}). At the higher level of Table~\ref{1111}, we evaluate the performance of different methods which use individual features for captioning prediction. Results indicate that using the long-range has the highest performance on all metrics. At the lower level of Table~\ref{1111}, we examine the impact of progressively combining different features together. Our full model using all three features outperforms all other counterparts.\\
\textbf{Different weighted metrics in seeding phase.}
The seeding phase training is important as it produces the entrance model for the following boosting phase. Hence, we evaluate the impact of using different weighting metric (a.k.a. \texttt{B@4}, \texttt{M},  \texttt{R}, and  \texttt{C}) in training (see Table~\ref{table:DXE_results}). On \texttt{R} and \texttt{C}, models trained by different DXE loss all outperform the counterpart trained by XE. Meanwhile, using CIDEr as the metric weight in DXE training obtain the best results on all metrics.
\begin{table}[tb]
\centering
\scalebox{0.83}{\begin{tabular}{c|cccc}
    \toprule 
    $b$  & B@4     & M  & R  & C   \\
    \midrule
    -  & 13.5 & 16.1 & 46.1 & 12.7 \\
    \midrule
    $b_{scst}$ & 44.6$\uparrow_{31.1}$ & 30.2$\uparrow_{14.1}$ & 64.3$\uparrow_{18.2}$ & 56.4$\uparrow_{43.7}$ \\
    Ours($b_1$)  &  46.4$\uparrow_{32.9}$    & 30.5$\uparrow_{14.4}$  & 65.0$\uparrow_{18.9}$  &  58.1$\uparrow_{45.4}$  \\
    Ours($b_2$) & \textbf{46.9$\uparrow_{33.4}$}   & \textbf{31.2$\uparrow_{15.1}$}  & \textbf{65.7$\uparrow_{19.6}$}   & \textbf{60.6$\uparrow_{47.9}$}  \\
	\bottomrule
\end{tabular}}
\caption{The performance of using different baselines as discrepant reward. $\uparrow$ indicates the increase from ``-'' (without baseline). We use DXE in the seeding phase.
}
\label{table:RL_baseline_results}
\end{table}\\
\textbf{Incremental training analysis.} We investigate if the incremental training could effectively improve our method performance. Table~\ref{table:germinating_from_XE_DXE} shows the results from the boosting phase increase steadily from the seeding phase. The boosting phase starting from the seeding phase using DXE gets higher scores than its counterpart using the XE. This proves that using DXE as the supervision in the seeding phase can yield more optimal model parameters and further improve the performance of the boosting phase than using XE in training.\\
\textbf{Using $b_1$ vs. $b_2$ in the boosting phase.} Table \ref{table:RL_baseline_results} shows the  results of models using different discrepant ($b_1$ and $b_2$) rewards compared with self-critical baseline ($b_{scst}$).
% Using the reward ($b_2$) based on top $Q$ sentences sampled by the model can help supervise models to achieve a better performance than using the reward ($b_1$) based on the ground-truth sentences. Both $b_1$ and $b_2$ supervised models outperform methods without using the discrepant reward.
Using the reward with $b_2$ baseline based on top $Q$ sentences sampled by the model can help our GL-RG to achieve a better performance than using the reward with $b_1$ baseline which is based on $G$ ground-truth sentences. Both models with $b_1$ or $b_2$ outperform methods without using the discrepant reward or with the self-critical reward ($b_{scst}$).
% \subsection{Generalization}
% To evaluate the generalization of our method, we craft our global-local encoder in some of the recent methods and leverage our DXE and DR to supervise their feature learning for captioning prediction.  Table~\ref{last_tab} demonstrates the improvements of each  method over their original implementations. 
\section{Conclusion}
Video captioning is an important research topic, which has various downstream applications. 
In this paper,  we propose a GL-RG framework for video captioning, which leverages the global-local vision representation to achieve fine-grained captioning on video contents with an incremental training strategy. Experimental results on two benchmarks demonstrate the effectiveness of our approach. In future, we plan to explore dynamic weighting scheme to capture the preferences on different granularities. We also plan to investigate integrating more multi-modal information.

\small{
\bibliographystyle{named}
\bibliography{ijcai22}}

\end{document}